\documentclass{article}

\usepackage{arxiv}
\usepackage{microtype}
\usepackage{graphicx}
\usepackage{subfigure}
\usepackage{amsmath}
\usepackage{amssymb}
\usepackage{multirow}
\usepackage{booktabs} 

\usepackage[utf8]{inputenc} 
\usepackage[T1]{fontenc}    
\usepackage{hyperref}       
\usepackage{url}            
\usepackage{booktabs}       
\usepackage{amsfonts}       
\usepackage{nicefrac}       
\usepackage{microtype}      
\usepackage{lipsum}
\usepackage{graphicx}
\graphicspath{ {./images/} }
\newcommand{\beginsupplement}{%
        \setcounter{table}{0}
        \renewcommand{\thetable}{S\arabic{table}}%
        \setcounter{figure}{0}
        \renewcommand{\thefigure}{S\arabic{figure}}%
     }

\title{Deep Bayesian Gaussian Processes for Uncertainty Estimation in Electronic Health Records}

\author{
 Yikuan Li \\
  Deep Medicine\\
  University of Oxford\\
  \texttt{yikuan.li@wrh.ox.ac.uk} \\
   \And
 Shishir Rao \\
  Deep Medicine\\
  University of Oxford\\
  \And
 Abdelaali Hassaine \\
  Deep Medicine\\
  University of Oxford\\
    \And
 Rema Ramakrishnan\\
  Deep Medicine\\
  University of Oxford\\
    \And
 Yajie Zhu \\
  Deep Medicine\\
  University of Oxford\\
    \And
 Dexter Canoy \\
  Deep Medicine\\
  University of Oxford\\
    \And
 Gholamreza Salimi-Khorshidi \\
  Deep Medicine\\
  University of Oxford\\
      \And
 Thomas Lukasiewicz  \\
  Department of Computer Science\\
  University of Oxford\\
      \And
 Kazem Rahimi \\
  Deep Medicine\\
  University of Oxford\\
}

\begin{document}
\maketitle
\begin{abstract}
One major impediment to the wider use of deep learning for clinical decision making is the difficulty of assigning a level of confidence to model predictions. Currently, deep Bayesian neural networks and sparse Gaussian processes are the main two scalable uncertainty estimation methods. However, deep Bayesian neural network suffers from lack of expressiveness, and more expressive models such as deep kernel learning, which is an extension of sparse Gaussian process, captures only the uncertainty from the higher level latent space. Therefore, the deep learning model under it lacks interpretability and ignores uncertainty from the raw data. In this paper, we merge features of the deep Bayesian learning framework with deep kernel learning to leverage the strengths of both methods for more comprehensive uncertainty estimation. Through a series of experiments on predicting the first incidence of heart failure, diabetes and depression applied to large-scale electronic medical records, we demonstrate that our method is better at capturing uncertainty than both Gaussian processes and deep Bayesian neural networks in terms of indicating data insufficiency and distinguishing true positive and false positive predictions, with a comparable generalisation performance. Furthermore, by assessing the accuracy and area under the receiver operating characteristic curve over the predictive probability, we show that our method is less susceptible to making overconfident predictions, especially for the minority class in imbalanced datasets. Finally, we demonstrate how uncertainty information derived by the model can inform risk factor analysis towards model interpretability.
\end{abstract}


\section{Introduction}
The application of deep learning to medicine has been growing over recent years. Much of research in this field has been focusing on estimating and improving “point predictions” in form of personalised risk scores for a given medical event in one’s future, as for instance reported in innovative deep learning models such as Deepr~\cite{nguyen2016mathtt}, RETAIN~\cite{choi2016retain} and Doctor AI~\cite{choi2016doctor}. However, point predictions – quite naturally – are prone to overconfidence. Considering the significant consequences of decision making in clinical practice that is guided by model predictions, quantifying the uncertainty of predictions is proving to be a key step in putting these models to practice in medicine.

In the last several years, a new subfield of deep learning, called probabilistic deep learning, has drawn wide interest to provide probabilistic predictions and uncertainty estimations at the same time. The most promising methods are Bayesian deep learning (BDL)~\cite{wang2016bayesian,zhang2018advances} and sparse Gaussian processes (GP)~\cite{liu2018gaussian,snelson2006sparse}. In BDL, by placing a distribution over each of the model weights instead of treating them as point values, the uncertainty of the weights can be passed layer by layer to eventually estimate the uncertainty in the predictions. However, this approach usually requires a compromise between model complexity and expressiveness of variational distributions. On the contrary, the GP model, as a non-parametric model, is more flexible and expressive than BDL. This advantage comes, however, at the expense of the need to store and process the data points for the covariance matrix. This usually takes cubic time $\mathcal{O}(n^3)$ to calculate the inversion and determinant of the covariance matrix for inference~\cite{burt2019rates} and becomes a challenge when working with large-scale datasets. The state-of-the-art solution to this challenge is to use a small number of pseudo-points, i.e., inducing points, to approximate the data points. This enables the covariance matrix to be approximated by a lower-rank representation~\cite{titsias2009variational,bui2017unifying}. Since the entire dataset is summarised by a small number of inducing points, this method is called sparse GP. \cite{wilson2016deep} upgraded this framework to be more flexible and scalable by implementing a deep architecture beneath the kernel function as a feature extractor, which is known as deep kernel learning (DKL). Although the deep architecture provides a significant boost in representational power, the framework can only capture the uncertainty in the higher-level latent space (after the deep architecture). This results in a lack of interpretability, and failure to capture the uncertainty in the deep architecture.

The core idea of this paper is to combine the strengths of both frameworks by merging the BDL concept with the DKL framework. We expect that this could retain  the expressiveness from GP while capturing the uncertainty during feature extraction, leading to: (1) a probabilistic feature representation for more robust inducing points and kernel training; and (2) a more comprehensive uncertainty estimation. Additionally, we investigate how the uncertainty information naturally contained in the Bayesian components can contribute to interpretable risk factor analysis in medical research.

\section{Background}
\subsection{Task description}
\begin{figure}[h]
\vskip 0.1in
\begin{center}
\centerline{\includegraphics[width=\columnwidth]{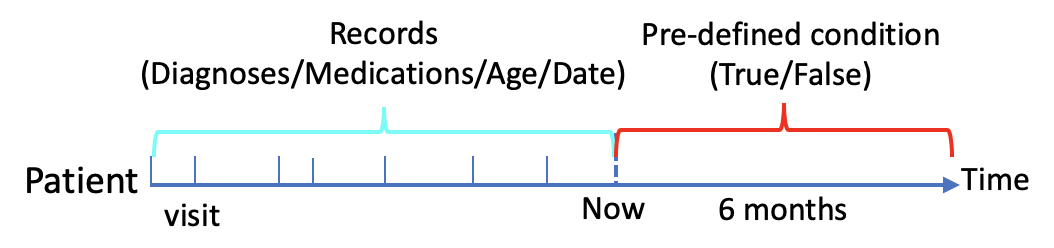}}
\caption{The axis represents the time; medical records before "now" are used to predict the incidence of a condition in the following six months; the medical records include multiple visits, and each visit has one or more diagnoses and medications; the interval between two visits represents the time duration between them.}
\label{fig:task}
\end{center}
\vskip -0.2in
\end{figure}

For this study, we explored a risk model for detecting the first incidence of three common chronic diseases, namely, heart failure (HF), diabetes and depression, using structured electronic health record (EHR) from the Clinical Practice Research Datalink (CPRD)~\cite{wolf2019data,rahimian2018predicting,herrett2015data}. We used diagnoses (ICD-10~\cite{icd10}), medications (British National Formulary code~\cite{bnf}), event date (time stamp for each diagnosis and medication) and date of birth as historical medical trajectory to predict whether the first incidence of aforementioned conditions would be diagnosed in the following six months for a patient, and the conditions are treated as separate prediction tasks. The design is summarised in Figure~\ref{fig:task}, and the ICD-codes for HF, diabetes, and depression are listed in Table~\ref{tab:conditions_hf},\ref{tab:conditions_diabetes},\ref{tab:conditions_depression}, in Supplementary~\ref{app:condition}. They were taken from previous publications~\cite{conrad2018temporal,kuan2019chronological}.

\subsection{Data source and cohort selection}\label{sec:cohort}
CPRD is one of the most comprehensive de-identified longitudinal population-based EHR datasets. It contains primary care data collected from a network of general practitioner practices across the UK, and it is also linked to Hospital Episode Statistics~\cite{herbert2017data} and other health and area-based administrative databases. The data encompass 42 million patients, including 13 million currently registered patients. The patients included are nationally representative in terms of age, sex, and ethnicity~\cite{herrett2015data}.

For this study, we set up a two-stage pipeline (A and B) for patient selection. Figure~\ref{fig:patient} in Supplementary~\ref{app:cohort} illustrates the procedures and the number of patients kept within each step. Stage A was a general data linkage step to select patients that met the minimum requirements for the study. This dataset was used for general unsupervised pre-training~\cite{lyu2018improving, miotto2016deep}. Stage B was designed for generating samples for the individual prediction tasks. Firstly, for a patient who had the pre-defined condition, records were formatted as in Figure~\ref{fig:task}, where all medical records before the first incidence of the condition were included as history records. For a patient who did not have the pre-defined condition, we randomly selected a time point to separate the records into history records and marked the patient as a negative sample. For all negative patients, we made sure that they had more than six months medical records after the selected time point to guarantee each of them was an absolute negative sample. Therefore, avoiding any assumption for the unseen future. The patient selection rules in stage B that kept patients who had enough records to be trained, eventually led to 788,880 (8.3\% positive samples), 913,799 (11.3\% positive samples) and 1,453,012 (16\% positive samples) patients for HF, diabetes and depression, respectively. We refer to the datasets from stages A and B as datasets A and B, respectively.

\section{Related work}
This section reviews the necessary concepts and related works.

\subsection{BEHRT}
\begin{figure}[h]
\vskip 0.1in
\begin{center}
\centerline{\includegraphics[width=\columnwidth]{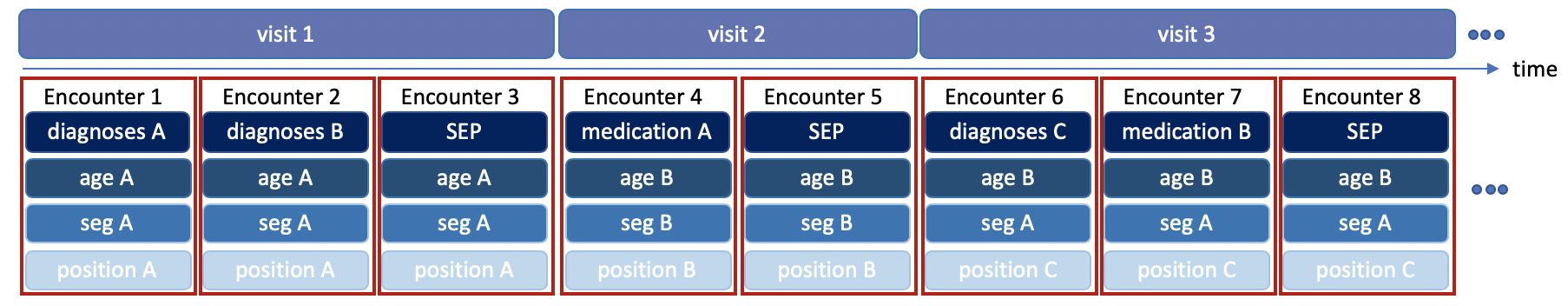}}
\caption{Four feature layers: clinical diagnoses and medications, age, segmentation and positional code; each visit could have multiple encounters and each encounter is a representation of multiple feature layers; summation of all embeddings are used for encounter representation.}
\label{fig:BEHRT}
\end{center}
\vskip -0.2in
\end{figure}

BEHRT~\cite{li2019behrt} is a recently developed model that applied the concept of self-attention transformer from natural language processing to EHRs. BEHRT took advantage of the self-attention mechanism and sequential format of EHRs in a way that maximally preserves the EHR-like structure. The feature structure is shown in Figure~\ref{fig:BEHRT}, with each encounter corresponding to the so-called "token" in transformers~\cite{devlin2018bert}. In this illustration, there are four embedding matrices for diagnoses and medications, age, segmentation and position separately. BEHRT uses the summation of the embeddings to represent each encounter. Recent work has shown that BEHRT outperform other deep learning models for disease prediction based in the context of complex large-scale sequential EHR. More detailed information can be found in~\cite{li2019behrt}. 

\subsection{Gaussian processes}
GPs are expressive non-parametric models~\cite{rasmussen2003gaussian}. We only consider regression at this stage, but they can be easily used for binary classification tasks by wrapping a logistic regression~\cite{murphy2012machine}. If we have observed training data, $\mathcal{D}=\{x_{i}, y_{i}\}^N_{i=1}$ with $x_{i} \in \chi$ and $y_{i} \in \mathbb{R}$, our target is to predict output $y^{*}$ for new inputs $x^{*}$. GPs usually place a GP prior over the latent function as $f \sim \mathcal{G}\mathcal{P}(v(\cdot), k(\cdot, \cdot))$, where $v:\chi\rightarrow{\mathbb{R}}$ is the mean function, and it is often taken as zero. The kernel function $k:\chi\times\chi\rightarrow{\mathbb{R}}$ controls the smoothness of GPs. A likelihood is then used to relate the latent function to the observed data through some noise, which is represented as $y_{i}=f(x_{i})+\epsilon_{i}$, $\epsilon_{i}\sim\mathcal{N}(0, \sigma^2_{n})$. In the end, we use the posterior for predictions and the marginal likelihood for selecting hyperparameters, which is shown in Equation~\ref{eq:gp}:
\begin{equation} \label{eq:gp}
log\,p(y)=-\frac{1}{2}\boldsymbol{y^T}\boldsymbol{K^{-1}_{n}}\boldsymbol{y}-\frac{1}{2}log|\boldsymbol{K_{n}}|-\frac{N}{2}log(2\pi)
\end{equation},
where $\boldsymbol{K_{n}}=\boldsymbol{K_{\text{f}\text{f}}}+\sigma^2_{n}\boldsymbol{I}$ and $\boldsymbol{|K_{\text{f}\text{f}}|}_{i,j}=k(x_{i},x_{j})$.

\subsection{Whitened-GPs}
Because of the complexity of calculating the determinant and inverse, many approaches have been proposed to approximate $\boldsymbol{K_{\text{f}\text{f}}}$ with a lower rank matrix~\cite{snelson2006sparse}. One popular approach is posterior approximation through variational free energy proposed by \cite{titsias2009variational}. The method suggests to optimise the evidence lower bound to minimise the KL divergence between posterior and variational distribution. Therefore, it directly approximats the posterior with a relatively small number of inducing points, and eventually simplifies the calculation. The evidence lower bound:
\begin{equation} \label{eq:elbo}
\begin{split}
\mathcal{L}_{SGP}=&-\frac{1}{2}\boldsymbol{y^T}\boldsymbol{Q^{-1}_{n}}\boldsymbol{y}-\frac{1}{2}log|\boldsymbol{Q_{n}}|-\frac{N}{2}log(2\pi)\\
&-\frac{t}{2\sigma^2_{n}}
\end{split}
\end{equation},
where $\boldsymbol{Q_{n}}=\boldsymbol{Q_{\text{f}\text{f}}}+\sigma^2_{n}\boldsymbol{I}$, $\boldsymbol{Q_{\text{f}\text{f}}}=\boldsymbol{K^T_{\text{u}\text{f}}K^{-1}_{\text{u}\text{u}}K_{\text{u}\text{f}}}$, $t=Tr(\boldsymbol{K_{\text{f}\text{f}}}-\boldsymbol{Q_{\text{f}\text{f}}})$, $\boldsymbol{[K_{\text{u}\text{f}}]}_{m,i}=k(z_m,x_i)$ and $\boldsymbol{[K_{\text{u}\text{u}}]}_{m,i}=k(u_m,u_i)$, $U=\{u_m\}^M_{m=1}$ represents inducing points; and SGP represents sparse GP.

\subsection{KISS-GPs}
Besides posterior approximation, \cite{wilson2015kernel} proposed a structured kernel interpolation framework to produce a more scalable kernel approximation, named KISS-GP. This method combines structure exploiting approaches, inducing points and sparse interpolation to further reduce the inference time cost and storage costs from $\mathcal{O}(m^2n+m^3)$ and $\mathcal{O}(mn+m^2)$, respectively, for whitened-GPs to $\mathcal{O}(n)$. Additionally, it provides a more accurate and a scalable inference and more flexible kernel learning on large datasets. The main idea of this method is to impose the grid constraint on the inducing points. Therefore, the kernel matrix $\boldsymbol{K_{\text{u}\text{u}}}$ admits the Kronecker structure for $\textit{d}>1$, where \textit{d} represents the dimension of the grid, The kernel could then be decomposed to a Kronecker product $k(u_{i},u_{j})=\prod^d_{t=1}k(u^t_{i}, u^t_{j})$. Additionally, if $\textit{d}=1$, the kernel is then a Toeplitz covariance matrix and can be calculated as $k(u_{i},u_{j})=k(u_{i}-u_{j})$. For the cross kernel matrix $\boldsymbol{K_{\text{u}\text{f}}}$, it is approximated, for example, by a local linear interpolation with adjacent grid inducing points; an example is as shown in Equation~\ref{eq:kiss-gp-kernel}:

\begin{equation} \label{eq:kiss-gp-kernel}
k(x_{i}, u_{j})\approx w_{i}k(u_{a}, u_{j})+(1-w_{i})k(u_{b},u_{j})
\end{equation},
where $u_{a}$ and $u_{b}$ are two inducing points on the grid that are closest to $x_{i}$, and $w_{i}$ is an interpolation weight that represents the distance to the inducing point. Eventually, the $\boldsymbol{Q_{\text{f}\text{f}}}$ matrix in Equation~\ref{eq:elbo} can be approximated as Equation~\ref{eq:kiss-gp}:

\begin{equation} \label{eq:kiss-gp}
\boldsymbol{Q_{\text{f}\text{f}}}\approx \boldsymbol{W}^T_{\text{u}\text{f}}\boldsymbol{K}^{-1}_{\text{u}\text{u}}\boldsymbol{W}_{\text{u}\text{f}}
\end{equation}

\subsection{Deep kernel learning}
In addition to sparse GP, \cite{wilson2016deep} moved one step further and proposed to embed deep neural networks (DNNs) with GPs to learn more flexible representations. The kernel function transforms from $k(x_{i},x_{j}|\boldsymbol{\theta})$ to $k(g(x_{i},\boldsymbol{w}),g(x_{j},\boldsymbol{w})|\boldsymbol{\theta},\boldsymbol{w})$, where $\boldsymbol{\theta}$ are the kernel hyperparameters, $g(\cdot)$ is a non-linear DNN, and $\boldsymbol{w}$ are the parametrised weights of the network. Therefore, the DNN acts as a feature extractor to represent samples as latent vectors, and GPs can make inferences based on the learned latent features.

\subsection{Variational inference for Bayesian deep learning}\label{sectionSVI}
Bayesian deep learning is another approach to implement a probabilistic model for uncertainty estimation. Instead of using point weights as deterministic DNNs, it places distributions over all model parameters~\cite{kingma2014adam}, and the predictive distribution can be estimated by marginalising the parameters~\cite{Gal2016Uncertainty,shridhar2019comprehensive}; this is shown as Equation~\ref{eq:predictive}.

\begin{equation} \label{eq:predictive}
p(y^*|x^*,\mathcal{D})=\int_\Omega p(y^*|x^*,\boldsymbol{w})p(\boldsymbol{w}|\mathcal{D})d\boldsymbol{w}
\end{equation},
where $\boldsymbol{w} \in \Omega$ represents parametrised weights. However, the posterior $p(\boldsymbol{w}|\mathcal{D})$ is usually intractable in neural networks. In order to retrieve it, \cite{blundell2015weight} proposed to approximate the posterior by optimising the evidence lower bound to minimise the KL divergence between variational distribution and posterior, which is shown in Equation~\ref{eq:elbo_bdl}: 

\begin{equation} \label{eq:elbo_bdl}
\begin{split}
\mathcal{L}_{BDL}=&\int_\Omega q_{\boldsymbol{\gamma}}(w)log(p(\mathcal{D}|\boldsymbol{w}))d\boldsymbol{w}\\
&-KL(q_{\boldsymbol{\gamma}}(\boldsymbol{w})||p(\boldsymbol{w}))
\end{split}
\end{equation},
where $q_{\boldsymbol{\gamma}}$ represents the variational distribution, which is parameterised by $\boldsymbol{\gamma}$. Afterwards, $p(\boldsymbol{w}|\mathcal{D})$ in Equation~\ref{eq:predictive} can be replaced by $q_{\boldsymbol{\gamma}}(\boldsymbol{w})$ for inference. 

\subsection{Challenges in uncertainty evaluation}\label{sec:uncertainty}
In recent research, high test log-likelihood has commonly been used to indicate the model's credibility to capture the true posterior~\cite{lakshminarayanan2017simple,gal2015dropout}. However, \cite{yao2019quality} did an experiment by constructing the "ground-truth" posterior and posterior predictive distribution by Hamiltonian Monte Carlo. They compared the approximated posteriors from inference methods such as probabilistic backpropagation, matrix-variate Gaussian and Bayes by hypernet with the "ground-truth". Even though the approximate posteriors incorrectly had a lower variance, they still yielded similar test log-likelihood as the ground-truth. Therefore, we argue that the test log-likelihood would be more meaningful to evaluate the posterior mean rather than the uncertainty (variance), and it is not a reliable criterium for determining how well an approximate posterior aligns with the true posterior. 

\section{Proposed methods}
\subsection{Deep Bayesian Gaussian processes}
In this section, we present the approach of our work, which combines the DKL with Bayesian inference for a more robust uncertainty estimation, and we refer to this architecture as deep Bayesian Gaussian processes (DBGPs). Additionally, we show how to learn the properties of these kernels as part of a scalable GP. We start with the kernel and inference of a GP, and the kernel hyperparameter $\boldsymbol{\theta}$ is ignored for the following parts to simplify the illustration. As for Equation~\ref{eq:kiss-gp-kernel}, the base kernel is shown as $k(x_{i}, x_{j})$; thus, the inference stage can be represented as:

\begin{equation}
p(\boldsymbol{f}^*) = \int p(\boldsymbol{f}^*|\boldsymbol{x}^*,\boldsymbol{f}_{m})p(\boldsymbol{f}_{m})d\boldsymbol{f}_m
\end{equation}

\begin{equation}
p(\boldsymbol{y}^*) = \int p(\boldsymbol{y}^*|\boldsymbol{f}^*)p(\boldsymbol{f}^*)d\boldsymbol{f}^*
\end{equation},
where $\boldsymbol{f}_m$ represents the latent prior from observed points or inducing points for GPs and sparse GPs, respectively. For DKL, the kernel is transformed to  $k(g(x_{i},\boldsymbol{w})$; $g(x_{j},\boldsymbol{w}))$, such a kernel function is used to measure the similarity between two latent representations extracted by a DNN.  Accordingly, the inference for $p(\boldsymbol{f}^*)$ is changed to Equation~\ref{eq:dkl-inference}, while $p(\boldsymbol{y}^*)$ remains the same.

\begin{equation}\label{eq:dkl-inference}
p(\boldsymbol{f}^*) = \int p(\boldsymbol{f}^*|\boldsymbol{x}^*, \boldsymbol{w},\boldsymbol{f}_{m})p(\boldsymbol{f}_{m})d\boldsymbol{f}_m
\end{equation}

In addition to the DKL, DBGP replaces the weights of the DNN as distributions. Therefore, besides marginalising the $\boldsymbol{f}_{m}$, we also need to marginalise the weights for the inference. The $p(\boldsymbol{f}^*)$ for DBGP is now transformed as follows:

\begin{equation}
p(\boldsymbol{f}^*) = \int_{\boldsymbol{\Omega}}\int p(\boldsymbol{f}^*|\boldsymbol{x}^*, \boldsymbol{w},\boldsymbol{f}_{m})p(\boldsymbol{w}|\mathcal{D})p(\boldsymbol{f}_{m})d\boldsymbol{f}_m d\boldsymbol{w}
\end{equation}

Since the posterior $p(\boldsymbol{w}|\mathcal{D})$ is usually intractable, we use a variational distribution $q_{\boldsymbol{\gamma}}(\boldsymbol{w})$ parametrised by $\boldsymbol{\gamma}$ to approximate it and then jointly train all the kernel hyperparameters \{$\boldsymbol{\theta}, \boldsymbol{\gamma}$\} together by optimising the evidence lower bound as Equation~\ref{eq:DBGP_ELBO}, and update the hyperparameters through chain rule in Equations~\ref{eq:update_kernel} and ~\ref{eq:update_model}:

\begin{equation}\label{eq:DBGP_ELBO}
\mathcal{L}_{DBGP}=\mathcal{L}_{SGP}-KL(q_{\boldsymbol{\gamma}}(\boldsymbol{w})||p(\boldsymbol{w}))
\end{equation}

\begin{equation}\label{eq:update_kernel}
\frac{\partial \mathcal{L}_{DBGP}}{\partial \boldsymbol{\theta}}=\frac{\partial \mathcal{L}_{DBGP}}{\partial \mathcal{L}_{SGP}}\frac{\partial \mathcal{L}_{SGP}}{\partial K}\frac{\partial K}{\partial \boldsymbol{\theta}}
\end{equation}

\begin{equation}\label{eq:update_model}
\begin{split}
\frac{\partial \mathcal{L}_{DBGP}}{\partial \boldsymbol{\gamma}}=&\frac{\partial \mathcal{L}_{DBGP}}{\partial \mathcal{L}_{SGP}}\frac{\partial \mathcal{L}_{SGP}}{\partial K}\frac{\partial K}{\partial q_{\boldsymbol{\gamma}}( \boldsymbol{w})}\frac{\partial{q_{\boldsymbol{\gamma}}( \boldsymbol{w})}}{\partial \boldsymbol{\gamma}}\\
&-\frac{\partial \mathcal{L}_{DBGP}}{\partial KL}\frac{\partial KL}{q_{\boldsymbol{\gamma}}( \boldsymbol{w})}\frac{\partial{q_{\boldsymbol{\gamma}}( \boldsymbol{w})}}{\partial \boldsymbol{\gamma}}
\end{split}
\end{equation}

\begin{figure}[h]
\vskip 0.1in
\begin{center}
\centerline{\includegraphics[width=\columnwidth]{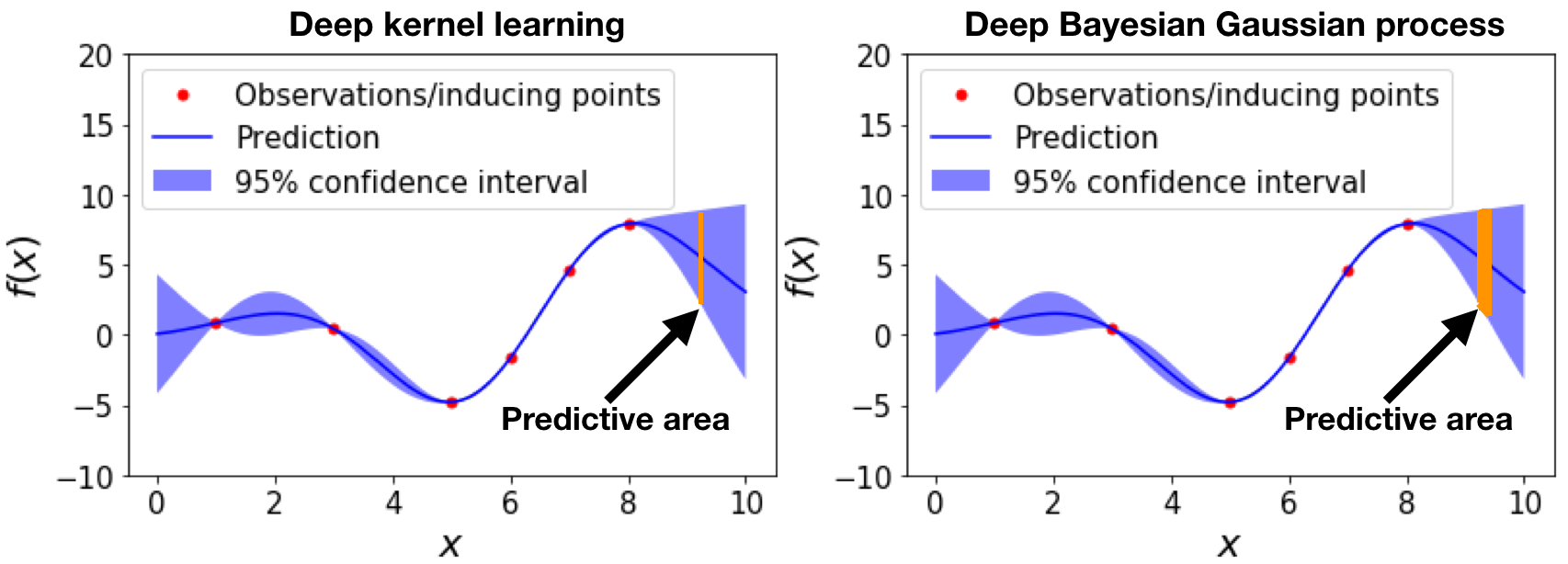}}
\caption{Predictive distribution for DKL and DBGP; the yellow area indicates the potential area for the predicted values.}
\label{fig:DBGP}
\end{center}
\vskip -0.2in
\end{figure}

Figure~\ref{fig:DBGP} provides an illustration of the potential difference between DKL and DBGP in terms of the predictive coverage. Unlike DKL, which maps the raw features into a fixed latent representation ($x$ in the figure) by a deep architecture, where uncertainty estimation of $f(x)$ completely comes from the GP, DBGP is able to capture the uncertainty hierarchically. The uncertainty captured by the deep Bayesian architecture is firstly reflected on the uncertainty in the latent representation ($x$ in the figure). Afterwards, such uncertainty in the latent representation moves forward to the GP. Eventually, the predictive uncertainty represents the uncertainty in both $x$ and $f(x)$ dimensions, which is more comprehensive and shows a wider predictive coverage. 

\subsection{Experiments}
\begin{figure}[h]
\vskip 0.1in
\begin{center}
\centerline{\includegraphics[width=\columnwidth]{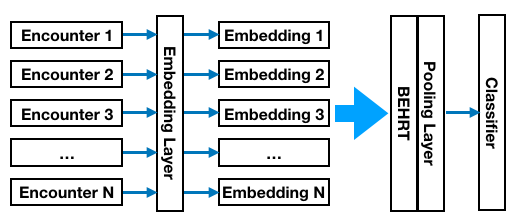}}
\caption{Model architecture; the embedding and the pooling are also included in BEHRT; the pooling layer only extracts the latent representation of the first encounter (time step) for classification.}
\label{fig:model}
\end{center}
\vskip -0.2in
\end{figure}

In our experiments, the model includes a feature extractor (BEHRT) and a classifier, as shown in Figure~\ref{fig:model}. We compare the proposed DBGP with DBL and DKL. The definitions of different architectures are listed below:\\
\textbf{Bayesian Embedding + KISS-GP (DBGP)}: The proposed DBGP framework with a Bayesian BEHRT as a feature extractor, in which the embedding parameters are stochastic, and a KISS-GP as the classifier; all the other parameters are deterministic.\\
\textbf{Bayesian Embeddings (BE)}: A Bayesian BEHRT in which the embedding parameters are stochastic, and all the other parameters are deterministic.\\
\textbf{Bayesian Output (BO)}: A Bayesian BEHRT in which the classifier parameters are stochastic, and all the other parameters are deterministic.\\
\textbf{Bayesian Embedding + Output (BE + BO)}: A Bayesian BEHRT in which the embedding and classifier parameters are stochastic, and all other parameters are deterministic.\\
\textbf{Whitened-GP}: A BEHRT model with a whitened-GP as the classifier, all the other parameters are deterministic.\\
\textbf{KISS-GP}: A BERHT model with a KISS-GP as the classifier, all the other parameters are deterministic.\\

\cite{dusenberry2019analyzing} indicates that for BDL, models with Bayesian embedding and Bayesian output usually work better than fully Bayesian models, so we mainly investigate the Bayesian model with stochastic embedding and output. Additionally, all the Bayesian components are formulated with a mean field distribution with zero mean for the prior, and all GP components are implemented with a multivariate distribution with zero mean and identity covariance matrix for the prior. The details of the other model parameters can be found in Table~\ref{model_params} in Supplementary~\ref{app:model}, which is selected by Bayesian hyperparameter optimisation.

\subsection{Evaluation methods}\label{sec:evalutaion}
In this work, Monte Carlo~\cite{Gal2016Uncertainty} was used for all probabilistic models to estimate the predictive distribution, and we evaluated the model performance from three perspectives: generalisation, ability of rejecting overconfident predictions and uncertainty estimation. For generalisation, we evaluated the area under the receiver operating characteristics (AUROC) curve and average precision (AP)~\cite{liu2009encyclopedia}; both of them were calculated based on the mean predictive probability. As for the ability of rejecting overconfident predictions, we evaluated the accuracy and AUROC as a function of the mean predictive probability. In medicine, it is highly desirable to avoid overconfident and incorrect predictions. Therefore, it is more useful to evaluate the model performance for predictions above a user-specified threshold. One tends to trust the predictions more when the confidence is high, and resorts to a different resolution when the prediction is not confident. Thus, the better rejection ability can be directly reflected by having a higher performance for high-confident predictions. For uncertainty estimation, we propose (1) to treat the uncertainty measurement for classification differently by measuring the difference between the variance of true positives and the variance of false positives. We would intuitively expect that the variance of true positives is distinguishably lower than the one of false positives. On the contrary, if the variance for both true positives and false positives are similar, we would say the model cannot provide a meaningful uncertainty estimation; (2) for imbalanced datasets, because the model is usually biased by the majority class, we would intuitively expect the model to have a higher uncertainty for the minority classes. Therefore, these two criteria are used as indication of the quality of the uncertainty estimation. To quantify the performance of the former criteria, we propose to calculate the Kullback-Leibler (KL) divergence between the uncertainty (standard deviation (std)) distribution of true positives and the uncertainty (std) distribution of false positives as Equation~\ref{eq:uncertainty_quantify}, and the larger the value is, the better the performance is. We can assume that both distributions are Gaussian distributions because of the central limit theorem~\cite{rosenblatt1956central}.

\begin{equation}\label{eq:uncertainty_quantify}
DIV = KL(p(FP)||p(TP))
\end{equation}

Here KL represents the KL divergence, $p(FP)$ is the distribution of false positives' uncertainty, and $p(TP)$ is the distribution of true positives' uncertainty. 

\subsection{Uncertainty analyses in embeddings}
In addition to the model performance evaluation, for models with Bayesian embeddings, we explore the linkage between embedding uncertainty and relative risk associations. For diagnoses or medications with high uncertainty in the embeddings, we assume they would contribute more uncertainty to the latent representations as well as the predictions. Therefore, higher uncertainty in the embedding indicates an unclear association between a disease or a medication and the target disease, otherwise, the association would be more certain even though the magnitude of the association is strong or weak. We used entropy~\cite{wang2008probability} to indicate the embedding uncertainty, and a detailed discussion will be covered in Section~\ref{sec:embeddingAnalysis}.

\section{Results}
We used the Monte Carlo method to sample each patient's prediction 30 times as an estimation of the predictive distribution for all analyses within this section. For comparison, we repeated the analysis with 60 samples and found no material difference in the prediction (Supplementary~\ref{app:results60}). 

\subsection{Generalisation performance}
\begin{table}[h!]
\vskip -0.1in
\caption{Prediction performance: metrics are calculated based on the mean predictive probability of 30 samples in the validation set.}
\label{tab:result}
\vskip 0.1in
\begin{center}
\begin{small}
\begin{sc}
\begin{tabular}{p{2cm}p{0.5cm}p{0.5cm}p{0.5cm}p{0.5cm}p{0.5cm}p{0.5cm}}
\toprule
& \multicolumn{2}{c}{HF} &\multicolumn{2}{c}{Diabetes}&\multicolumn{2}{c}{Depression}\\
\midrule
Model & A & B & A & B & A & B \\
\midrule
\textbf{DBGP} & 0.941 & 0.625 & 0.834 & 0.533 & 0.776 & 0.416\\
Whitened GP & 0.945 & 0.645 & 0.835 & 0.540 & 0.782 & 0.432\\
KISS-GP & 0.945 & 0.650 & 0.836 & 0.536 & 0.782 & 0.433\\
BE & 0.942& 0.630 & 0.831 & 0.529 & 0.774 & 0.409\\
BO & 0.932 & 0.644 & 0.824 & 0.524 & 0.767 & 0.426\\
BE+BO & 0.941 & 0.628 & 0.835 & 0.537 & 0.778 & 0.419\\
\midrule
\multicolumn{7}{c}{A: AUROC, B: Average precision}\\
\bottomrule
\end{tabular}
\end{sc}
\end{small}
\end{center}
\vskip -0.1in
\end{table}

We compared the performance for generalisation of aforementioned six probabilistic models. In our experiments, we split each dataset (HF, diabetes and depression) into 70\% training set and 30\% validation set. Before training for a specific prediction task, we pre-trained the original deterministic BEHRT model on the dataset A, which is explained in Section~\ref{sec:cohort}, based on an unsupervised masked language model task~\cite{devlin2018bert}. Then, to fine-tune the model for the prediction task, we initialised the deterministic parameters with the pre-trained model parameters, and for all stochastic components, we used the pre-trained parameters as the mean of their variational distribution. Since the classifier was not a part of the masked language model training task, all the parameters within the classifier were randomly initialised. Table~\ref{tab:result} demonstrates the performance for the marginalised prediction performance of each probabilistic model. It shows that all implemented probabilistic models have a comparable performance in terms of generalisation. The result is expected, because all models share the same fundamental BEHRT model.

\subsection{Accuracy and AUROC as a function of confidence}
We re-used the results from the experiments in the previous section to evaluate the ability of rejecting overconfident predictions based on the mean predictive probability.

\begin{figure}[h]
\vskip 0.1in
\begin{center}
\centerline{\includegraphics[width=\columnwidth]{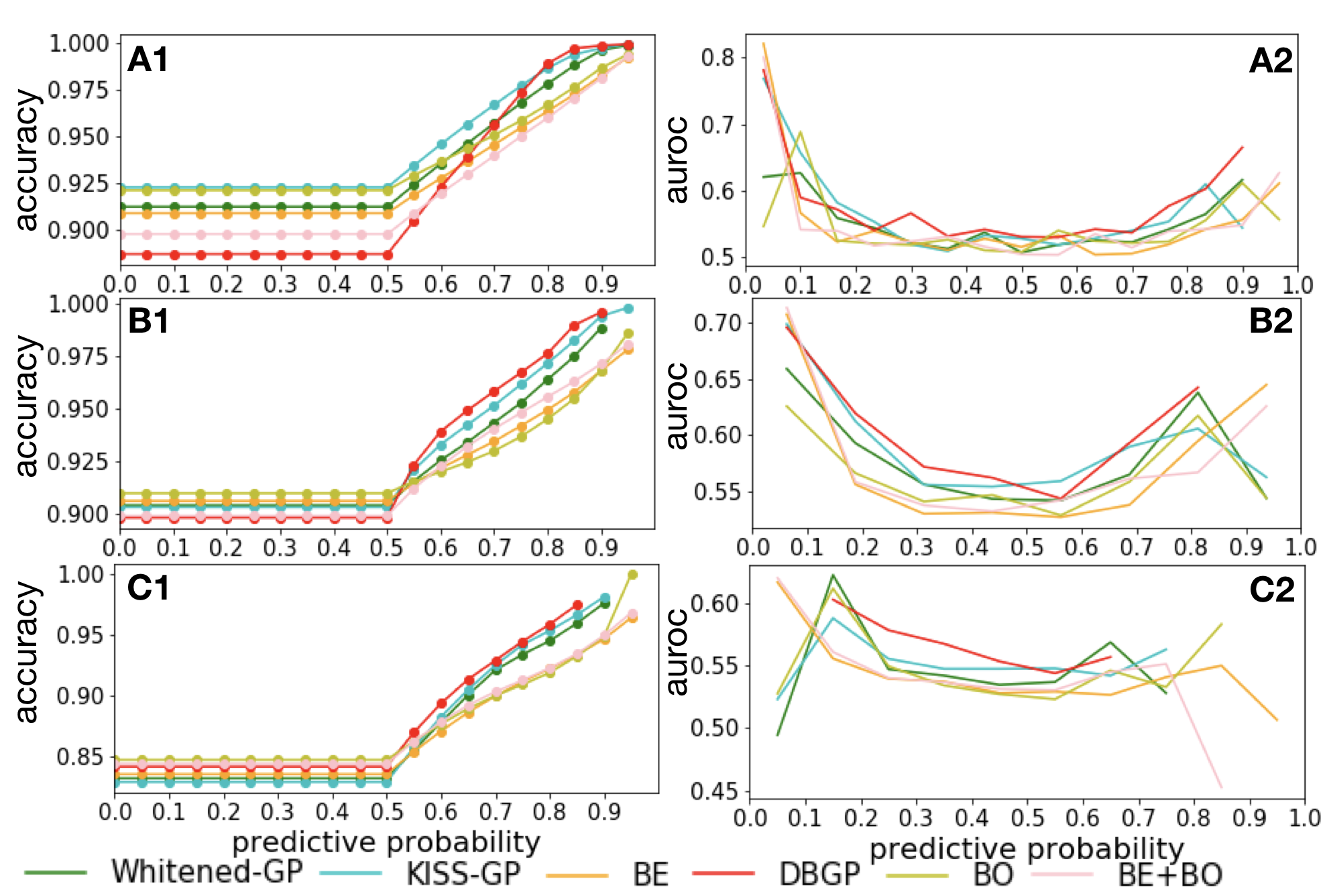}}
\caption{Accuracy and AUROC vs confidence curves; A: Heart failure, B: Diabetes, C: Depression; the x-axis represents the mean predictive probability, and the y-axis represents the accuracy or AUROC.}
\label{fig:auroc_bin}
\end{center}
\vskip -0.2in
\end{figure}

For the accuracy vs confidence curve, we treated the prediction as a two class classification task, with the mean prediction $p(y=k|x)$, where $k$ represents the $k^{th}$ class, which is 2 in total in our case (binary classification). We defined the predicted label as $\hat{y} = argmax_{k}p(y=k|x)$ and the confidence as $p(y=\hat{y}|x)=max_{k}p(y=k|x)$. The performance of accuracy for patients with confidence above different thresholds is shown in Figure~\ref{fig:auroc_bin}. A1, B1 and C1 are the evaluations for HF, diabetes and depression respectively. Because we only considered samples with confidence above each threshold, we would expect a rise in accuracy with increase in confidence thresholds. The figures show that the GP-based methods outperform the BDL models. Furthermore, the proposed method DBGP shows a better performance than the other models, especially for the high-confident predictions.

Additionally, unlike the measurement for the accuracy vs confidence curve, to know more details about the model performance over different predictive probabilities, we used the mean predictive probability carried out by the models to evaluate the performance over the predictive confidence in terms of AUROC. The results are shown in A2, B2 and C2 in Figure~\ref{fig:auroc_bin}. These figures illustrate that the proposed DBGP model has a better AUROC over the predictive confidence in general. Furthermore, it shows more robust predictions for the minority (positive) class. Instead of giving overconfident predictions, DBGP shows a better capability of penalising highly confident predictions, and such effects become more significant when the model prediction performance drops (among HF, diabetes and depression prediction). 

\subsection{Uncertainty estimation}
In this experiment, we evaluated the performance of uncertainty estimation for models based on the aforementioned two criteria in Section~\ref{sec:evalutaion}: (1) predictions should intuitively have a higher uncertainty for the minority class than the majority class in the imbalanced dataset; and (2) the true positive predictions should have less uncertainty than the false positive predictions. 

\begin{figure}[h]
\vskip 0.1in
\begin{center}
\centerline{\includegraphics[width=\columnwidth]{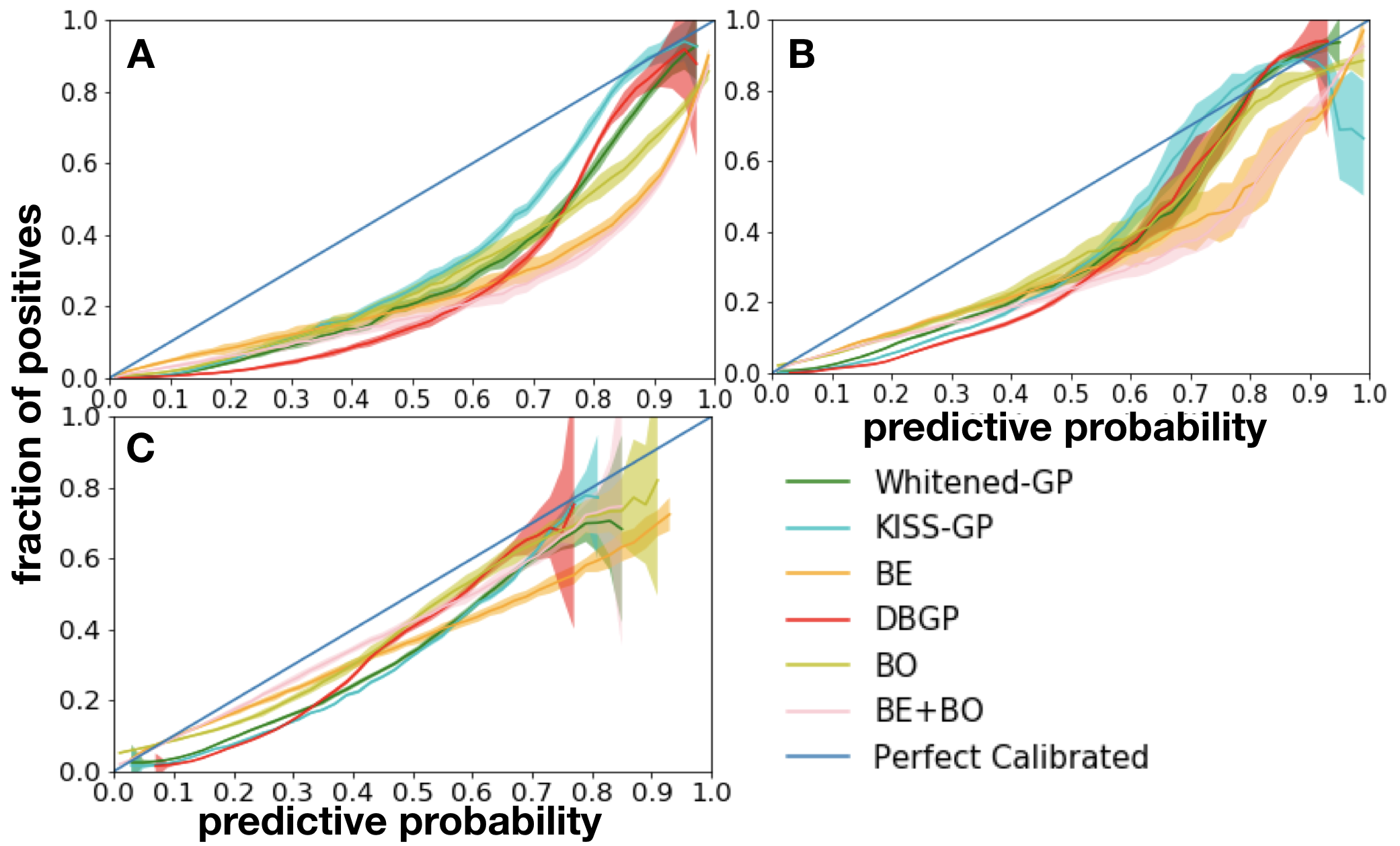}}
\caption{Calibration curve; the x and the y axis represent the predictive probability and the fraction of positive predictions respectively; A: heart failure, B: diabetes, C: depression}
\label{fig:calibration}
\end{center}
\vskip -0.2in
\end{figure}

To evaluate the uncertainty across different predictive probabilities, instead of using the mean predictive probability, we calculated the empirical frequency across the predictive probability for each set of samples. Therefore, we can estimate the uncertainty of predictions over predictive probabilities. The results are shown in Figure~\ref{fig:calibration}. 

The figure shows the consistency with the previous results in Figure~\ref{fig:auroc_bin} that the proposed method DBGP has a better rejection capability for the biased positive predictions, and the rejection becomes significant when the generalisation performance drops. Additionally, it also indicates that GP-based methods are better at capturing uncertainties for positive predictions in imbalanced datasets than the BDL-based models. For BDL-based models, we can observe that for rhe HF prediction task, even though the prediction performance in terms of average precision is relatively poor, they are still very certain for those high-confidence positive predictions. The uncertainty only starts to show when the model performance is extremely poor, as shown in depression prediction task. In contrast, DBGP shows the capability of capturing the uncertainty for the minority class, and also illustrates a better calibration for positive predictions. 

\begin{figure}[h]
\vskip 0.1in
\begin{center}
\centerline{\includegraphics[width=\columnwidth]{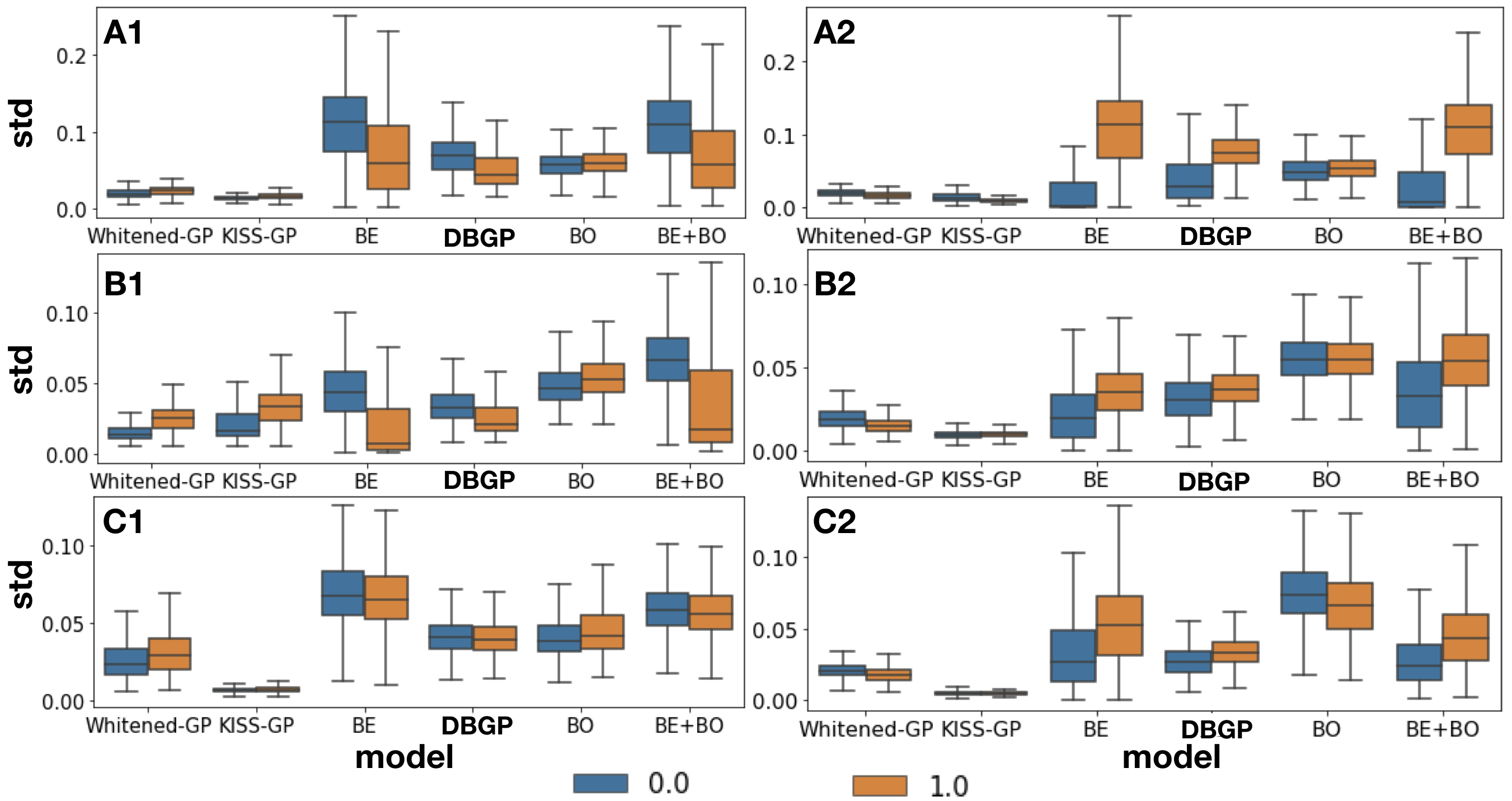}}
\caption{Distribution of std for positive and negative predictions; A, B and C represent the heart failure, diabetes and depression prediction tasks, respectively; 1 represents samples with mean predictive probability larger than 0.5 (positive prediction), and 2 represents samples with mean predictive probability less than 0.5 (negative prediction), for instance: A1 means positive prediction for HF prediction task; orange and blue represent samples with label positive and negative, respectively.}
\label{fig:uncertainty_pos_neg}
\end{center}
\vskip -0.2in
\end{figure}

Additionally, to evaluate the uncertainty difference between true positives and false positives, we represented each patient with mean predictive probability and corresponding std calculated from samples from predictive distributions. Furthermore, we considered samples with predictive probability higher than 0.5 to be predicted as positives and predictive probability less than 0.5 to be predicted as negatives. Figure~\ref{fig:uncertainty_pos_neg} uses the boxplot to show the distribution of std for all positive predictions and negative predictions. It shows that the Bayesian-embedding-based methods have better a capability to capture the uncertainty to distinguish the true positives and false positives, true negatives and false negatives (for the convenience of description, we also consider true negative as true positive and false negative as false positive). On the contrary, the GP-based methods either provide an indistinguishable uncertainty estimation for true positives and false positives, or provide uncertainty estimations that seem incorrect, as shown by the uncertainty for true positives being even higher than the false positives. Therefore, for those models that show the correct trend, we propose to calculate the divergence between the distribution of true positives and the distribution of false positives to quantify the quality of distinguishability, as explained in Section~\ref{sec:evalutaion}. The results are shown in Table~\ref{tab:divergence}. 

\begin{table}[h!]
\caption{Uncertainty divergence (DIV): DIV calculates the KL divergence between the std distributions of false positive and true positive predictions.}
\label{tab:divergence}
\vskip 0.1in
\begin{center}
\begin{small}
\begin{sc}
\begin{tabular}{p{1.5cm}p{0.5cm}p{0.5cm}p{0.5cm}p{0.5cm}p{0.5cm}p{0.5cm}}
\toprule
& \multicolumn{2}{c}{HF} &\multicolumn{2}{c}{Diabetes}&\multicolumn{2}{c}{Depression}\\
\midrule
Model & P & N & P & N & P & N \\
\midrule
\textbf{DBGP} & 0.330 & 0.867 & 0.105 & 0.146 & 0.009 & 0.279\\
BE & 0.298& 0.995 & 0.585 & 0.286 & 0.009 & 0.250\\
BE+BO & 0.360 & 0.224 & 0.427 & 0.311 & 0.009 & 0.158\\
\midrule
\multicolumn{7}{c}{P: Positive prediction, N: Negative Prediction}\\
\bottomrule
\end{tabular}
\end{sc}
\end{small}
\end{center}
\vskip -0.1in
\end{table}

Table~\ref{tab:divergence} indicates that the uncertainty estimation is ambiguous for true positives and false positives if the model prediction performance is relatively poor, such as for depression. As the prediction performance improves (diabetes), the Bayesian methods become slightly better than DBGP. However, when the prediction performance reaches a certain level (HF), DBGP equally performs as well as the deep Bayesian models.

\subsection{Embedding analysis}\label{sec:embeddingAnalysis}
The DBGP framework not only improves the ability of uncertainty estimation, but also brings a certain level of interpretability to the underlying deep architecture. In this section, we analysed the uncertainty of Bayesian embeddings by measuring its entropy for the DBGP model to understand how the embedding affects the uncertainty in the latent representation. Entropy is a commonly used metric to quantify the uncertainty of a probability distribution~\cite{wang2008probability}.

Because we only imposed the uncertainty to the embedding layer, all the uncertainty of the latent representation came from the embedding. Intuitively, a higher uncertainty of the embedding contributes more to the uncertainty of the latent representation. Therefore, if an embedding has a high uncertainty, it can mean a diagnosis or medication has more complex contextual information, and its association with the prediction is more unclear. In this case, such information can give us guidance to: (1) whether a causal or direct association is more likely to be included in the low uncertainty (low entropy) group; (2) whether the joint association has a higher chance to be included in the high uncertainty (high entropy) group. We list the diagnoses and medications with top 10 lowest entropy from HF in Table~\ref{tab:entropy_low}, more information about entropy values for diabetes and depression can be found in Table~\ref{tab:entropy_low_rest} in Supplementary~\ref{app:entropy} and the Table~\ref{tab:entropy_high} of Supplementary~\ref{app:entropy} illustrates the diagnoses and medications with highest entropy.

\begin{table}[h!]
\caption{Entropy measurement for diagnoses and medications; top 10 lowest entropy diagnoses and medications for \textbf{heart failure}. The entropy is calculated by the summation of the entropy of the distribution over all embedding dimensions.}
\label{tab:entropy_low}
\begin{center}
\begin{small}
\begin{sc}
\begin{tabular}{lp{6cm}}
\toprule
Entropy & description\\
\midrule
-236.14 & Diuretics\\
-235.67 & Acute myocardial infarction, unspecified\\
-235.43 & Unspecified acute lower respiratory infection\\
-235.17 & Other ill-defined heart diseases\\
-235.12 & Antiplatelet Drugs\\
-235.08 & Chronic obstructive pulmonary disease with (acute) exacerbation\\
-234.97 & Chronic obstructive pulmonary disease, unspecified\\
-234.95 & Torticollis\\
-234.90 & Cellulitis and acute lymphangitis of other parts of limb\\
-234.68 & Type 2 diabetes mellitus with circulatory complications\\
\bottomrule
\end{tabular}
\end{sc}
\end{small}
\end{center}
\vskip -0.1in
\end{table}

Table~\ref{tab:entropy_low} shows that most of the low entropy diagnoses and medications are closely associated to HF, and similar patterns can also be found in diabetes and depression. This method suggest a link between uncertainty and risk factor analyses and can potentially be used to further assist the model interpretability and guide future causality analyses.

\section{Discussion}
In this paper, we have proposed a mixed architecture named DBGP. It combines the strengths from both GPs and BDL, showing strong performance for rejecting overconfident predictions and providing a more robust uncertainty estimation. From the experiments in this paper, we have noticed that our methods can provide more reasonable estimation for the true positive and false positive predictions. Additionally, it could indicate the insufficiency and the robustness for biased predictions in the imbalanced dataset, whereas GPs and BDL can only show promising results for one of those two aspects. Furthermore, we investigated the associations between uncertainty and risk factors, and the results showed strong evidence for the relation. Therefore, an interesting topic for future work lies in an interpretability and causality analysis.

\section{Acknowledgements}
This research was funded by the Oxford Martin School (OMS) and supported by the National Institute for Health Research (NIHR) Oxford Biomedical Research Centre (BRC). The views expressed are those of the authors and not necessarily those of the OMS, the UK National Health Service (NHS), the NIHR or the Department of Health and Social Care. This work uses data provided by patients and collected by the NHS as part of their care and support and would not have been possible without access to this data. The NIHR recognises and values the role of patient data, securely accessed and stored, both in underpinning and leading to improvements in research and care.

\bibliographystyle{unsrt}  
\bibliography{references}  

\begin{thebibliography}{10}

\bibitem{nguyen2016mathtt}
Phuoc Nguyen, Truyen Tran, Nilmini Wickramasinghe, and Svetha Venkatesh.
\newblock {Deepr: A Convolutional Net for Medical Records}.
\newblock {\em IEEE journal of biomedical and health informatics},
  21(1):22--30, 2016.

\bibitem{choi2016retain}
Edward Choi, Mohammad~Taha Bahadori, Jimeng Sun, Joshua Kulas, Andy Schuetz,
  and Walter Stewart.
\newblock {RETAIN: An Interpretable Predictive Model for Healthcare using
  Reverse Time Attention Mechanism}.
\newblock In {\em Advances in Neural Information Processing Systems}, pages
  3504--3512, 2016.

\bibitem{choi2016doctor}
Edward Choi, Mohammad~Taha Bahadori, Andy Schuetz, Walter~F Stewart, and Jimeng
  Sun.
\newblock {Doctor AI: Predicting Clinical Events via Recurrent Neural
  Networks}.
\newblock In {\em Machine Learning for Healthcare Conference}, pages 301--318,
  2016.

\bibitem{wang2016bayesian}
Hao Wang and Dit-Yan Yeung.
\newblock {Towards Bayesian Deep Learning: A Survey}, 2016.

\bibitem{zhang2018advances}
Cheng Zhang, Judith Butepage, Hedvig Kjellstrom, and Stephan Mandt.
\newblock {Advances in Variational Inference}.
\newblock {\em IEEE transactions on pattern analysis and machine intelligence},
  2018.

\bibitem{liu2018gaussian}
Haitao Liu, Yew-Soon Ong, Xiaobo Shen, and Jianfei Cai.
\newblock {When Gaussian Process Meets Big Data: A Review of Scalable GPs}.
\newblock {\em arXiv preprint arXiv:1807.01065}, 2018.

\bibitem{snelson2006sparse}
Edward Snelson and Zoubin Ghahramani.
\newblock {Sparse Gaussian Processes using Pseudo-inputs}.
\newblock In {\em Advances in neural information processing systems}, pages
  1257--1264, 2006.

\bibitem{burt2019rates}
David~R Burt, Carl~E Rasmussen, and Mark Van Der~Wilk.
\newblock {Rates of Convergence for Sparse Variational Gaussian Process
  Regression}.
\newblock {\em arXiv preprint arXiv:1903.03571}, 2019.

\bibitem{titsias2009variational}
Michalis Titsias.
\newblock {Variational Learning of Inducing Variables in Sparse Gaussian
  Processes}.
\newblock In {\em Artificial Intelligence and Statistics}, pages 567--574,
  2009.

\bibitem{bui2017unifying}
Thang~D Bui, Josiah Yan, and Richard~E Turner.
\newblock {A Unifying Framework for Gaussian Process Pseudo-Point
  Approximations using Power Expectation Propagation}.
\newblock {\em The Journal of Machine Learning Research}, 18(1):3649--3720,
  2017.

\bibitem{wilson2016deep}
Andrew~Gordon Wilson, Zhiting Hu, Ruslan Salakhutdinov, and Eric~P Xing.
\newblock {Deep Kernel Learning}.
\newblock In {\em Artificial Intelligence and Statistics}, pages 370--378,
  2016.

\bibitem{wolf2019data}
Achim Wolf, Daniel Dedman, Jennifer Campbell, Helen Booth, Darren Lunn,
  Jennifer Chapman, and Puja Myles.
\newblock {Data resource profile: Clinical Practice Research Datalink (CPRD)
  Aurum}.
\newblock {\em International journal of epidemiology}, 2019.

\bibitem{rahimian2018predicting}
Fatemeh Rahimian, Gholamreza Salimi-Khorshidi, Amir~H Payberah, Jenny Tran,
  Roberto~Ayala Solares, Francesca Raimondi, Milad Nazarzadeh, Dexter Canoy,
  and Kazem Rahimi.
\newblock {Predicting the risk of emergency admission with machine learning:
  Development and validation using linked electronic health records}.
\newblock {\em PLoS medicine}, 15(11):e1002695, 2018.

\bibitem{herrett2015data}
Emily Herrett, Arlene~M Gallagher, Krishnan Bhaskaran, Harriet Forbes, Rohini
  Mathur, Tjeerd van Staa, and Liam Smeeth.
\newblock {Data resource profile: Clinical Practice Research Datalink (CPRD)}.
\newblock {\em International journal of epidemiology}, 44(3):827--836, 2015.

\bibitem{icd10}
Robin~L Walker, Deirdre~A Hennessy, Helen Johansen, Christie Sambell, Lisa Lix,
  and Hude Quan.
\newblock {Implementation of ICD-10 in Canada: how has it impacted coded
  hospital discharge data?}
\newblock {\em BMC health services research}, 12(1):149, 2012.

\bibitem{bnf}
OL~Wade and GD~McDevitt.
\newblock {Prescribing and the British national formulary}.
\newblock {\em British medical journal}, 2(5514):635, 1966.

\bibitem{conrad2018temporal}
Nathalie Conrad, Andrew Judge, Jenny Tran, Hamid Mohseni, Deborah Hedgecott,
  Abel~Perez Crespillo, Moira Allison, Harry Hemingway, John~G Cleland, John~JV
  McMurray, et~al.
\newblock {Temporal trends and patterns in heart failure incidence: a
  population-based study of 4 million individuals}.
\newblock {\em The Lancet}, 391(10120):572--580, 2018.

\bibitem{kuan2019chronological}
Valerie Kuan, Spiros Denaxas, Arturo Gonzalez-Izquierdo, Kenan Direk, Osman
  Bhatti, Shanaz Husain, Shailen Sutaria, Melanie Hingorani, Dorothea Nitsch,
  Constantinos~A Parisinos, et~al.
\newblock {A chronological map of 308 physical and mental health conditions
  from 4 million individuals in the English National Health Service}.
\newblock {\em The Lancet Digital Health}, 2019.

\bibitem{herbert2017data}
Annie Herbert, Linda Wijlaars, Ania Zylbersztejn, David Cromwell, and Pia
  Hardelid.
\newblock {Data Resource Profile: Hospital Episode Statistics Admitted Patient
  Care (HES APC)}.
\newblock {\em International journal of epidemiology}, 46(4):1093--1093i, 2017.

\bibitem{lyu2018improving}
Xinrui Lyu, Matthias Huser, Stephanie~L Hyland, George Zerveas, and Gunnar
  Ratsch.
\newblock {Improving Clinical Predictions through Unsupervised Time Series
  Representation Learning}.
\newblock {\em arXiv preprint arXiv:1812.00490}, 2018.

\bibitem{miotto2016deep}
Riccardo Miotto, Li~Li, Brian~A Kidd, and Joel~T Dudley.
\newblock {Deep Patient: An Unsupervised Representation to Predict the Future
  of Patients from the Electronic Health Records}.
\newblock {\em Scientific reports}, 6:26094, 2016.

\bibitem{li2019behrt}
Yikuan Li, Shishir Rao, Jose Roberto~Ayala Solares, Abdelaali Hassaine, Dexter
  Canoy, Yajie Zhu, Kazem Rahimi, and Gholamreza Salimi-Khorshidi.
\newblock {BEHRT: Transformer for Electronic Health Records}, 2019.

\bibitem{devlin2018bert}
Jacob Devlin, Ming-Wei Chang, Kenton Lee, and Kristina Toutanova.
\newblock {BERT: Pre-training of Deep Bidirectional Transformers for Language
  Understanding}, 2018.

\bibitem{rasmussen2003gaussian}
Carl~Edward Rasmussen.
\newblock {Gaussian Processes in Machine Learning}.
\newblock In {\em Summer School on Machine Learning}, pages 63--71. Springer,
  2003.

\bibitem{murphy2012machine}
Kevin~P Murphy.
\newblock {\em {Machine Learning: A Probabilistic Perspective}}.
\newblock 2012.

\bibitem{wilson2015kernel}
Andrew Wilson and Hannes Nickisch.
\newblock {Kernel Interpolation for Scalable Structured Gaussian Processes
  (KISS-GP)}.
\newblock In {\em International Conference on Machine Learning}, pages
  1775--1784, 2015.

\bibitem{kingma2014adam}
Diederik~P Kingma and Jimmy Ba.
\newblock {Adam: A Method for Stochastic Optimization}.
\newblock {\em arXiv preprint arXiv:1412.6980}, 2014.

\bibitem{Gal2016Uncertainty}
Yarin Gal.
\newblock {\em {Uncertainty in Deep Learning}}.
\newblock PhD thesis, University of Cambridge, 2016.

\bibitem{shridhar2019comprehensive}
Kumar Shridhar, Felix Laumann, and Marcus Liwicki.
\newblock {A Comprehensive guide to Bayesian Convolutional Neural Network with
  Variational Inference}.
\newblock {\em arXiv preprint arXiv:1901.02731}, 2019.

\bibitem{blundell2015weight}
Charles Blundell, Julien Cornebise, Koray Kavukcuoglu, and Daan Wierstra.
\newblock {Weight Uncertainty in Neural Networks}, 2015.

\bibitem{lakshminarayanan2017simple}
Balaji Lakshminarayanan, Alexander Pritzel, and Charles Blundell.
\newblock {Simple and Scalable Predictive Uncertainty Estimation using Deep
  Ensembles}.
\newblock In {\em Advances in Neural Information Processing Systems}, pages
  6402--6413, 2017.

\bibitem{gal2015dropout}
Yarin Gal and Zoubin Ghahramani.
\newblock {Dropout as a Bayesian Approximation: Representing Model Uncertainty
  in Deep Learning}, 2015.

\bibitem{yao2019quality}
Jiayu Yao, Weiwei Pan, Soumya Ghosh, and Finale Doshi-Velez.
\newblock {Quality of Uncertainty Quantification for Bayesian Neural Network
  Inference}.
\newblock {\em arXiv preprint arXiv:1906.09686}, 2019.

\bibitem{dusenberry2019analyzing}
Michael~W. Dusenberry, Dustin Tran, Edward Choi, Jonas Kemp, Jeremy Nixon,
  Ghassen Jerfel, Katherine Heller, and Andrew~M. Dai.
\newblock {Analyzing the Role of Model Uncertainty for Electronic Health
  Records}, 2019.

\bibitem{liu2009encyclopedia}
Ling Liu and M~Tamer {\"O}zsu.
\newblock {\em {Encyclopedia of database systems}}, volume~6.
\newblock Springer New York, NY, USA:, 2009.

\bibitem{rosenblatt1956central}
Murray Rosenblatt.
\newblock {A Central Limit Theorem and a Strong Mixing Condition}.
\newblock {\em Proceedings of the National Academy of Sciences of the United
  States of America}, 42(1):43, 1956.

\bibitem{wang2008probability}
Qiuping~A Wang.
\newblock {Probability distribution and entropy as a measure of uncertainty}.
\newblock {\em Journal of Physics A: Mathematical and Theoretical},
  41(6):065004, 2008.

\end{thebibliography}

\newpage
\appendix
\beginsupplement

\section*{Supplementary}
\section{Condition description}\label{app:condition}

\begin{table}[h!]
\caption{Heart failure conditions}
\label{tab:conditions_hf}
\begin{center}
\begin{tabular}{lp{6cm}}
\toprule
ICD-10 & Description \\
\midrule
I09.9 & Rheumatic heart disease, unspecified\\
I11.0 & Hypertensive heart disease with (congestive) heart failure\\
I13.0 & Hypertensive heart and renal disease with (congestive) heart failure \\
I13.2 & Hypertensive heart and chronic kidney disease with heart failure and with stage 5 chronic kidney disease, or end stage renal disease\\
I25.5 & Ischaemic cardiomyopathy \\
I27.9 & Pulmonary heart disease, unspecified\\
I38 & Endocarditis, valve unspecified\\
I42.0 & Dilated cardiomyopathy\\
I42.1 & Obstructive hypertrophic cardiomyopathy\\
I42.2 & Other hypertrophic cardiomyopathy\\
I42.6 & Alcoholic cardiomyopathy\\
I42.8 & Other cardiomyopathies\\
I42.9 & Cardiomyopathy, unspecified\\
I50.0 & Congestive heart failure\\
I50.1 & Left ventricular failure\\
I50.2 & Systolic (congestive) heart failure\\
I50.3 & Diastolic (congestive) heart failure\\
I50.8 & Other heart failure\\
I50.9 & Heart failure, unspecified\\
\bottomrule
\end{tabular}
\end{center}
\end{table}

\begin{table}[h]
\caption{Diabetes conditions}
\label{tab:conditions_diabetes}
\begin{center}
\begin{tabular}{lp{6cm}}
\toprule
ICD-10 & Description \\
\midrule
E10.0-E10.9 & Type 1 diabetes mellitus\\
E11.0-E11.9 & Type 2 diabetes mellitus\\
E12.0-E12.9 & Malnutrition-related diabetes mellitus\\
E13.0-E13.9 & Other specified diabetes mellitus\\
E14.0-E14.9 & Unspecified diabetes mellitus\\
O24.2 & Pre-existing malnutrition-related diabetes mellitus\\
\bottomrule
\end{tabular}
\end{center}
\end{table}

\newpage
\begin{table}[ht]
\caption{Depression conditions}
\label{tab:conditions_depression}
\begin{center}
\begin{tabular}{lp{6cm}}
\toprule
ICD-10 & Description \\
\midrule
F32.0 & Mild depressive episode\\
F32.1 & Moderate depressive episode\\
F32.2 & Severe depressive episode without psychotic symptoms\\
F32.3 & Severe depressive episode with psychotic symptoms\\
F32.8 & Other depressive episodes\\
F32.9 & Depressive episode, unspecified\\
F33.0 & Recurrent depressive disorder, current episode mild\\
F33.1 & Recurrent depressive disorder, current episode mild\\
F33.2 & Recurrent depressive disorder, current episode severe without psychotic symptoms\\
F33.3 & Recurrent depressive disorder, current episode severe with psychotic symptoms\\
F33.4 & Recurrent depressive disorder, currently in remission\\
F33.8 & Other recurrent depressive disorders\\
F33.9 & Recurrent depressive disorder, unspecified\\
F34.1 & Dysthymia\\
F38.1 & Other recurrent mood [affective] disorders\\
\bottomrule
\end{tabular}
\end{center}
\end{table}

\section{Cohort selection}\label{app:cohort}
\begin{figure}[h!]
\vskip 0.1in
\begin{center}
\centerline{\includegraphics[width=\columnwidth]{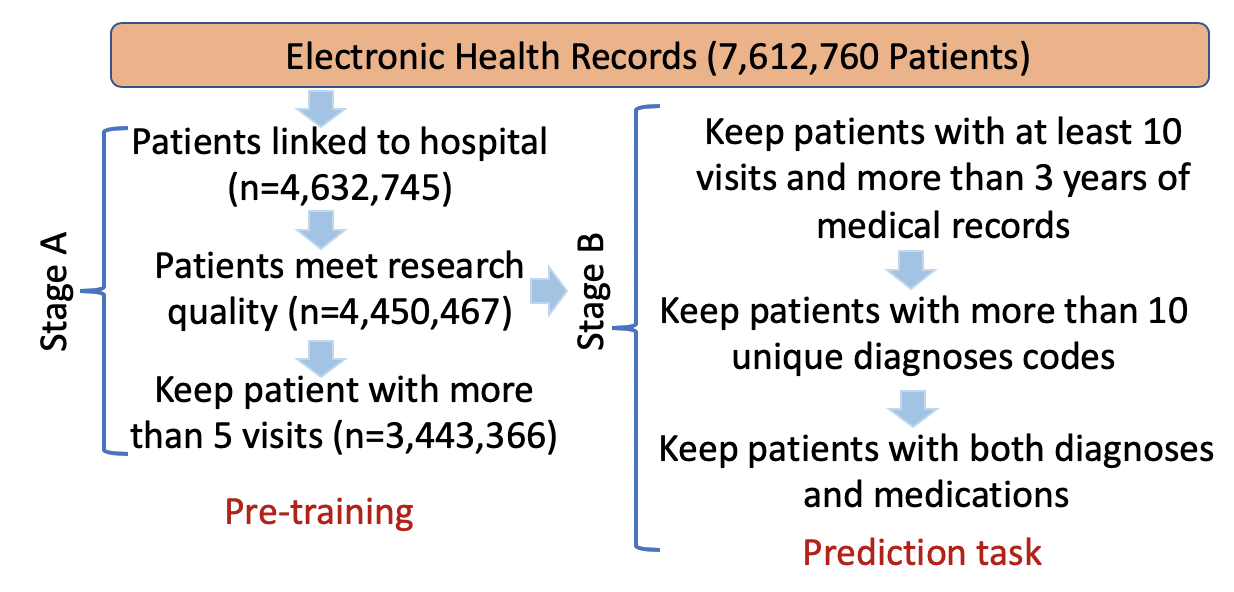}}
\caption{Patient selection; stage A illustrates the data cleaning pipeline from raw CPRD dataset to the dataset for model pre-training; and Stage B is used for patient selection for the incidence prediction tasks; the number of patients been kept in each step is represented as n.}
\label{fig:patient}
\end{center}
\vskip -0.2in
\end{figure}

\section{Model configuration}\label{app:model}
\begin{table}[h]
\caption{Model parameters}
\label{model_params}
\begin{center}
\begin{small}
\begin{tabular}{lccccccr}
\toprule
\multicolumn{8}{c}{All}\\
\midrule
Maximum Sequence Length & \multicolumn{7}{c}{256}\\
Hidden Size &\multicolumn{7}{c}{150}\\
Number of Hidden Layers &\multicolumn{7}{c}{4}\\
Number of Attention Heads &\multicolumn{7}{c}{6}\\
Intermediate Size &\multicolumn{7}{c}{108}\\
Hidden Dropout &\multicolumn{7}{c}{0.29}\\
\midrule
\multicolumn{2}{c}{}&\multicolumn{3}{c}{BE/BO/BE+BO}&\multicolumn{3}{c}{Others}\\
\midrule
\multicolumn{2}{c}{Pool Layer Size}& \multicolumn{3}{c}{150}&\multicolumn{3}{c}{24}\\
\midrule
\multicolumn{3}{c}{}&\multicolumn{3}{c}{BE/DBGP}&\multicolumn{1}{c}{BO/BE+BO}&\multicolumn{1}{c}{}\\
\midrule
\multicolumn{3}{c}{Embedding Prior Std} & \multicolumn{3}{c}{0.374}&\multicolumn{1}{c}{0.374}&\multicolumn{1}{c}{}\\
\multicolumn{3}{c}{Output Prior Std} & \multicolumn{3}{c}{-}&\multicolumn{1}{c}{1}&\multicolumn{1}{c}{}\\

\bottomrule
\end{tabular}
\end{small}
\end{center}
\end{table}

\newpage
\section{Entropy measurement}\label{app:entropy}
\begin{table}[h!]
\vskip 0.1in
\caption{Entropy measurement for diagnoses and medications; top 10 \textbf{lowest} entropy diagnoses and medications for \textbf{diabetes} and \textbf{depression}. Entropy is calculated by the summation of entropy of distribution over all embedding dimensions.}
\label{tab:entropy_low_rest}
\begin{center}
\begin{small}
\begin{sc}
\begin{tabular}{lp{6cm}}
\toprule
Entropy & description\\
\midrule
\multicolumn{2}{c}{Diabetes}\\
\midrule
-230.52 & Drugs Used In Diabetes\\
-230.28 & Detection Strips, Urine For Glycosuria\\
-229.18 & Unspecified diabetes mellitus\\
-229.16 & Hypoglycemia, unspecified\\
-229.00 & Obesity, unspecified\\
-228.42 & Essential (Primary) Hypertension\\
-227.24 & Positive Inotropic Drugs\\
-226.98 & Lipid-Regulating Drugs\\
-226.96 & Chronic tubulo-interstitial nephritis, unspecified\\
-226.77 & Sex Hormones\\
\midrule
\multicolumn{2}{c}{Depression}\\
\midrule
-229.68 & Antidepressant Drugs\\
-221.35 & Anxiety Disorder, Unspecified\\
-220.69 & Influenza due to unidentified influenza virus with other respiratory manifestations\\
-219.65 & Drugs Used In Psychoses and Rel.Disorders\\
-219.58 & Analgesics\\
-219.36 & Hypnotics And Anxiolytics\\
-219.24 & Laxatives\\
-219.15 & low back pain\\
-219.13 & General Anaesthesia/Hypnotics And Anxiolytics\\
-219.09 &  Dyspep and Gastro-Oesophageal Reflux Disease\\
\bottomrule
\end{tabular}
\end{sc}
\end{small}
\end{center}
\vskip -0.1in
\end{table}

\newpage
\begin{table}[h!]
\caption{Entropy measurement for diagnoses and medications; top 10 \textbf{highest} entropy diagnoses and medications for heart failure, diabetes and depression. Entropy is calculated by the summation of entropy of distribution over all embedding dimensions.}
\label{tab:entropy_high}
\begin{center}
\begin{small}
\begin{sc}
\begin{tabular}{lp{6cm}}
\toprule
Entropy & description\\
\midrule
\multicolumn{2}{c}{Heart failure} \\
\midrule
-226.61 & Traumatic subdural hemorrhage\\
-227.24 & Peritonsillar abscess\\
-227.39 & Cough Suppressants/Opioid Analgesics/Drugs Used In Substance Dependence - Opioid Dependence\\
-227.61 & Malignant neoplasm of main bronchus\\
-227.78 & Other specified noninflammatory disorders of uterus\\
-227.99 & Glaucoma suspect\\
-228.07 & Urostomy Bag\\
-228.13 & Synovitis and tenosynovitis, unspecified\\
-228.25 & Subjective visual disturbances\\
-228.27 & Delivery by elective caesarean section\\
\midrule
\multicolumn{2}{c}{Diabetes}\\
\midrule
-218.64 & Retained placenta without hemorrhage\\
-218.64 & Disease of stomach and duodenum, unspecified\\
-218.90 & Intraductal carcinoma in situ of breast\\
-218.97 & Acute noninfective otitis externa\\
-219.39 & Nontoxic goiter, unspecified\\
-219.49 & Malignant neoplasm of unspecified ovary\\
-219.54 & Drugs Used In Rheumatic Diseases and Gout/Preparations For Eczema And Psoriasis\\
-219.58 & Lateral epicondylitis\\
-219.71 & Delusional disorders\\
-219.97 & Dorsopathy, unspecified\\
\midrule
\multicolumn{2}{c}{Depression}\\
\midrule
-210.28 & Other specified disorders of synovium and tendon\\
-210.79 &Antihist, Hyposensit and Allergic Emergen/Cough Preparations\\
-210.81 & Viral wart, unspecified\\
-210.84 & Other specified viral infections characterized by skin and mucous membrane lesions\\
-210.95 & Candidiasis of skin and nail\\
-211.13 & Malignant neoplasm of head of pancreas\\
-211.16 & Mumps without complication\\
-211.24 & Dysthymia\\
-211.25 & Acute posthemorrhagic anemia\\
-211.25 & Fracture of upper end of tibia\\

\bottomrule
\end{tabular}
\end{sc}
\end{small}
\end{center}
\end{table}

\section{Results for 60 times of sampling from predictive distribution}\label{app:results60}

\subsection{Generalisation performance}
\begin{table}[h!]
\caption{Prediction performance: $N=60$ samples are sampled from the predictive distribution of each patient in the validation set for all probabilistic models, and both metrics are calculated based on the mean predictive probability of $N$ samples.}
\begin{center}
\begin{small}
\begin{sc}
\begin{tabular}{p{2cm}p{0.5cm}p{0.5cm}p{0.5cm}p{0.5cm}p{0.5cm}p{0.5cm}}
\toprule
& \multicolumn{2}{c}{HF} &\multicolumn{2}{c}{Diabetes}&\multicolumn{2}{c}{Depression}\\
\midrule
Model & A & B & A & B & A & B \\
\midrule
Whitened GP & 0.945 & 0.646 & 0.835 & 0.540 & 0.782 & 0.434\\
KISS-GP & 0.945 & 0.650 & 0.836 & 0.536 & 0.782 & 0.433\\
BE & 0.942& 0.631 & 0.831 & 0.529 & 0.774 & 0.410\\
DBGP & 0.942 & 0.627 & 0.834 & 0.534 & 0.776 & 0.416\\
BO & 0.935 & 0.648 & 0.828 & 0.526 & 0.772 & 0.428\\
BE+BO & 0.941 & 0.629 & 0.836 & 0.537 & 0.779 & 0.420\\
\midrule
\multicolumn{7}{c}{A: AUROC, B: Average precision}\\
\bottomrule
\end{tabular}
\end{sc}
\end{small}
\end{center}
\end{table}

\subsection{Accuracy and AUROC as a function of confidence}
\begin{figure}[h]
\begin{center}
\centerline{\includegraphics[width=\columnwidth]{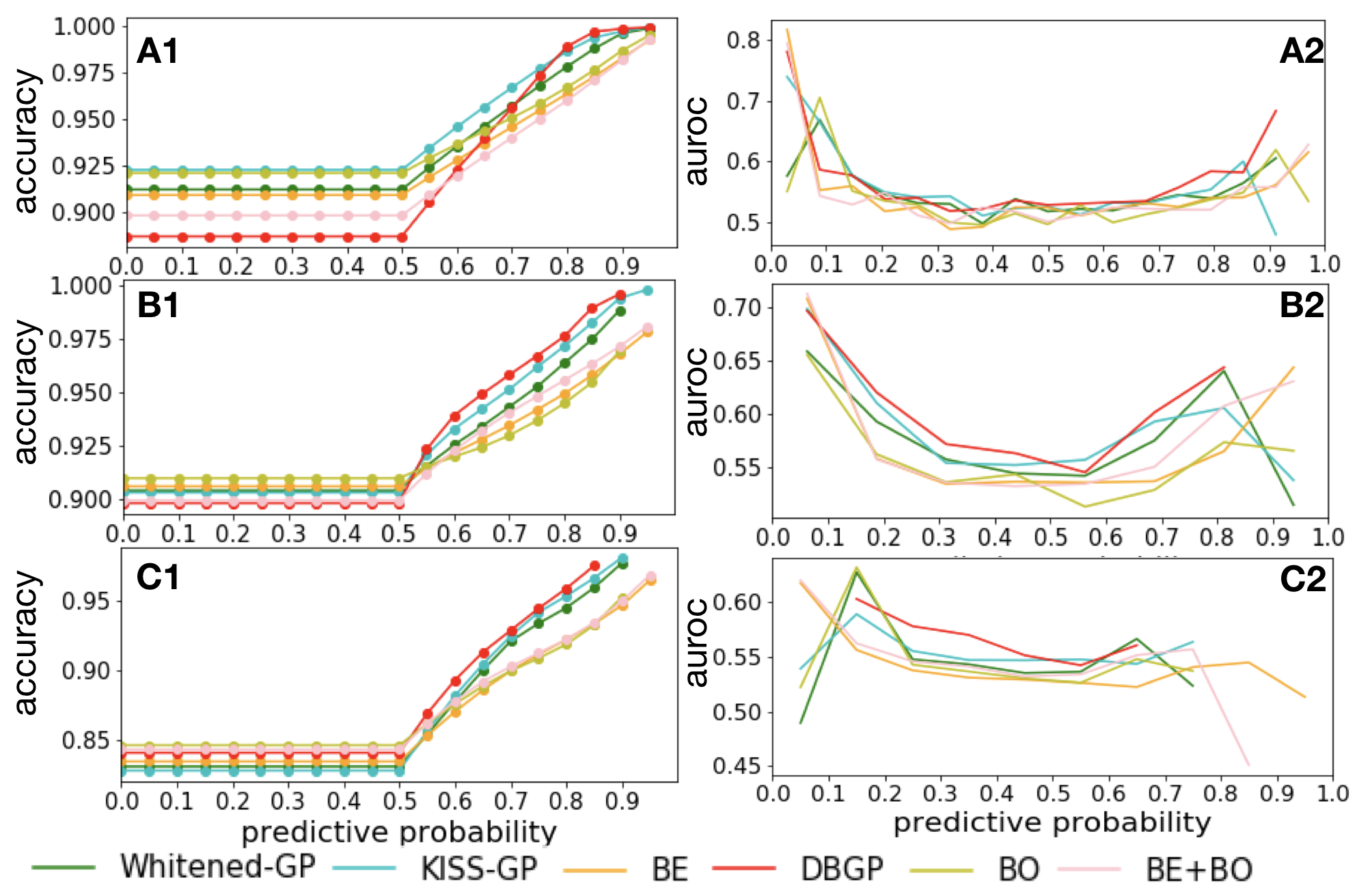}}
\caption{Accuracy and AUROC vs Confidence curves; N=60 samples are sampled from the predictive distribution of each patient, and metrics are calculated based on the mean predictive probability of N samples; A: Heart failure, B: Diabetes, C: Depression; x-axis represents the mean predictive probability and y-axis represents accuracy or auroc.}
\end{center}
\end{figure}

\newpage
\subsection{Uncertainty estimation}
\begin{figure}[h!]
\begin{center}
\centerline{\includegraphics[width=\columnwidth]{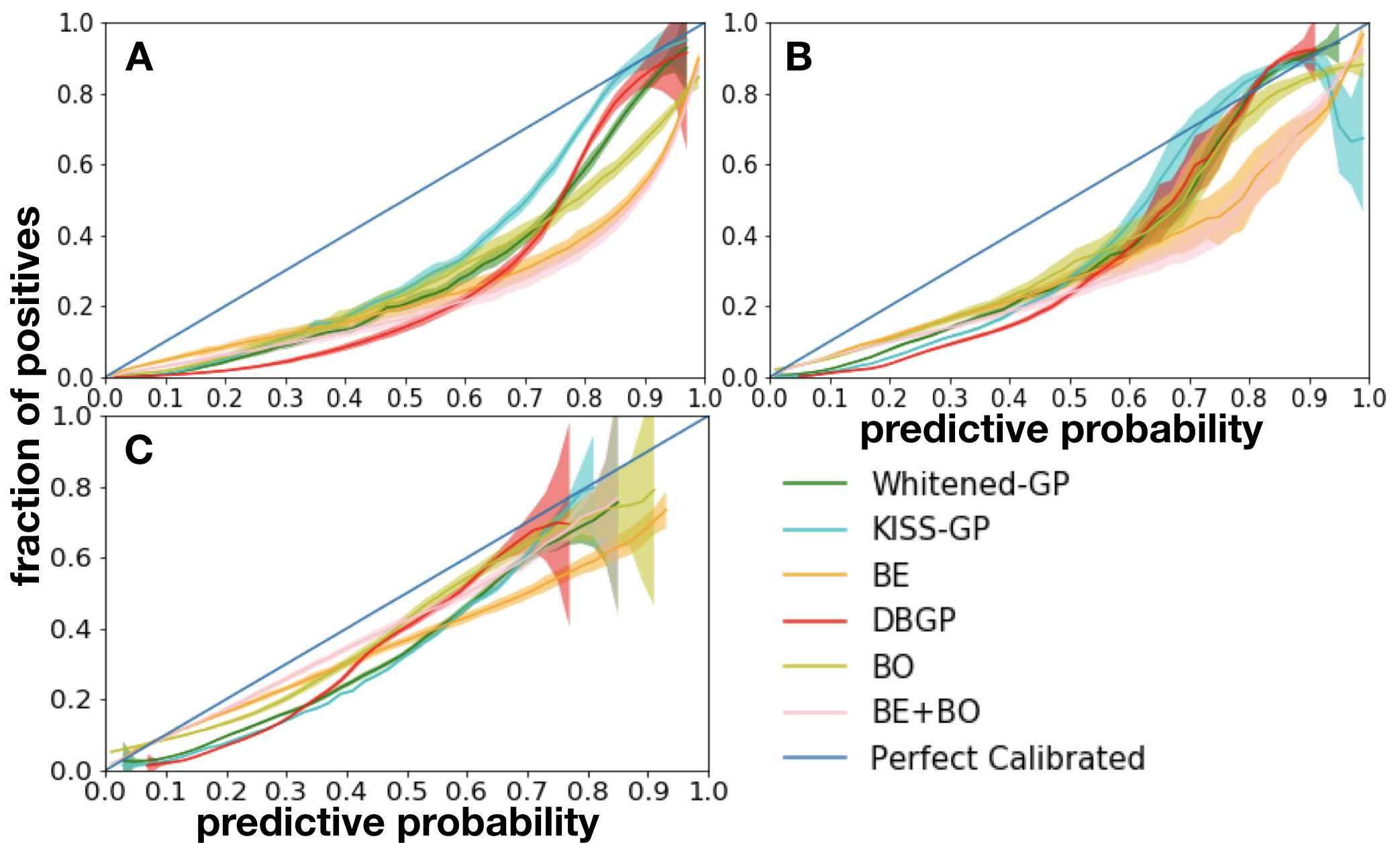}}
\caption{Calibration curve; x-axis represents the predictive probability and y-axis represents the empirical frequency; N=60 samples are sampled from the predictive distribution of each patient; A: heart failure, B: diabetes, C: depression.}
\end{center}
\end{figure}

\begin{figure}[h!]
\begin{center}
\centerline{\includegraphics[width=\columnwidth]{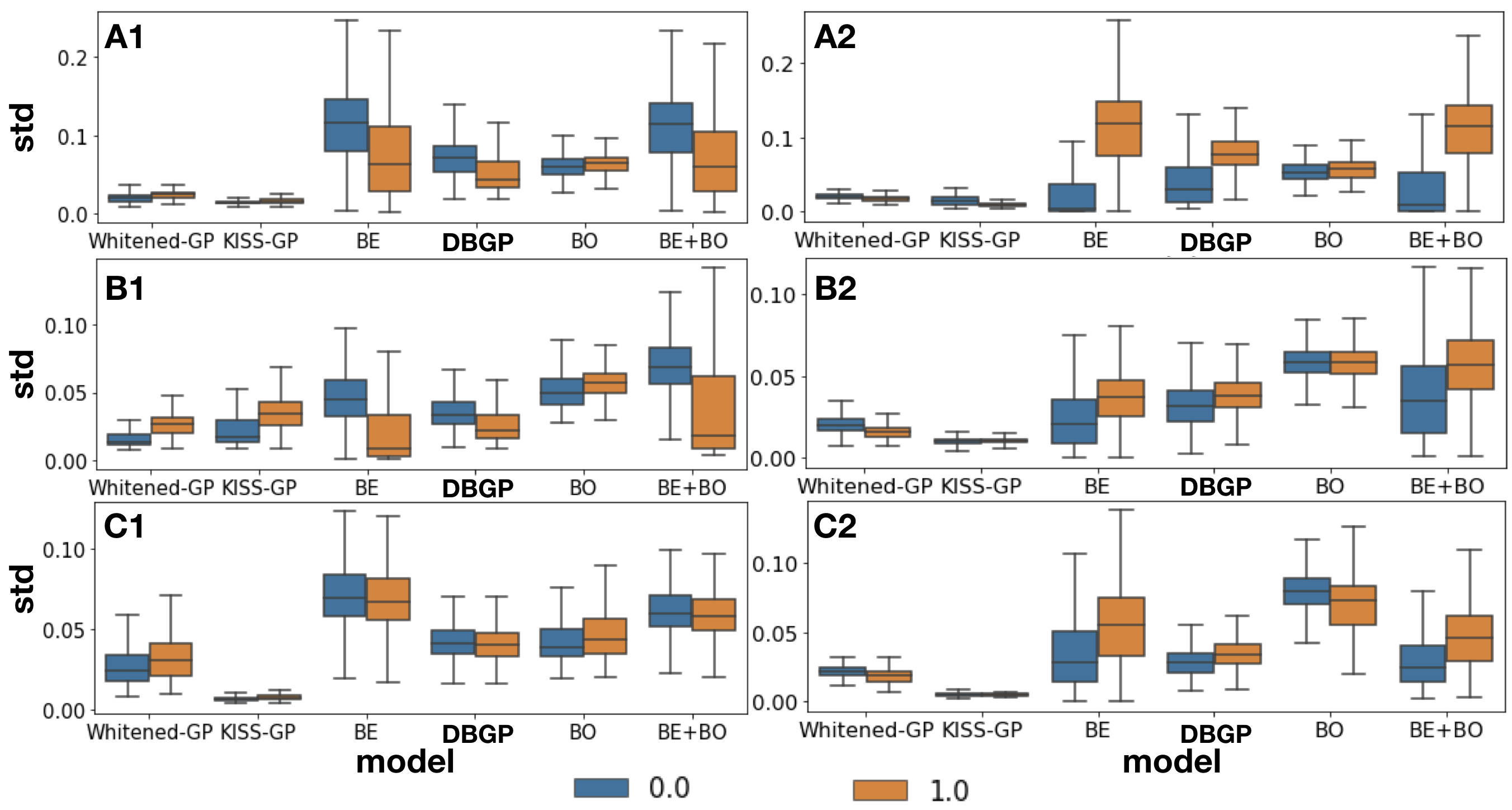}}
\caption{Distribution of std for positive and negative predictions; A, B and C represent heart failure, diabetes and depression prediction tasks respectively; 1 represents samples with mean predictive probability larger than 0.5 (positive prediction), and 2 represents samples with mean predictive probability less than 0.5 (negative prediction), for instance: A1 means positive prediction for heart failure prediction task; orange represents samples with label positive, and blue represents samples with label negative; the std for each patient is calculated from 60 times of sampling of a patient's predictive distribution.}
\end{center}
\end{figure}

\newpage
\section{population statistics for accuracy analysis}
\begin{figure}[h!]
\begin{center}
\centerline{\includegraphics[width=\columnwidth]{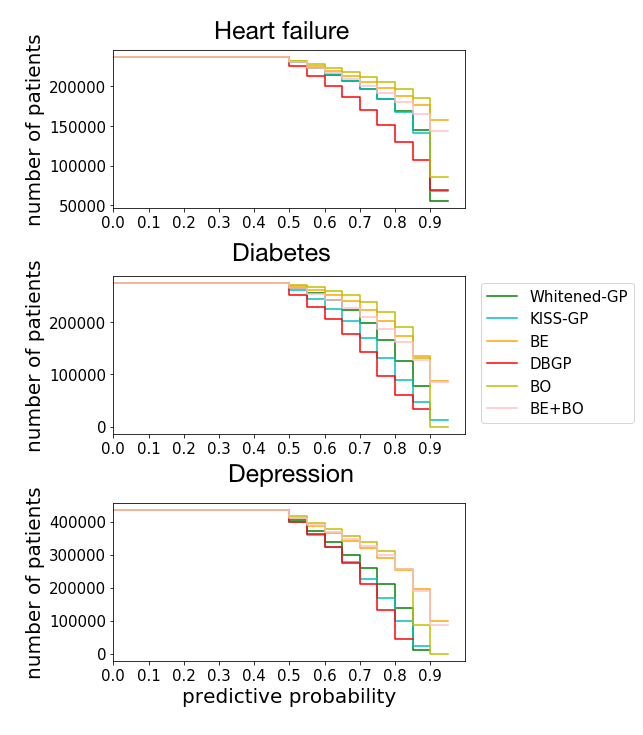}}
\caption{Number of patients above certain confidence threshold.}
\label{app:num_patient}
\label{fig:acc_pop_stats}
\end{center}
\end{figure}

Figure~\ref{fig:auroc_bin} shows the accuracy over different confidence threshold, here we present more details about the number of patient above each threshold in Figure~\ref{fig:acc_pop_stats}.

\end{document}